# A SAM based Tool for Semi-Automatic Food Annotation

**Lubnaa Abdur Rahman, Ioannis Papathanail, Lorenzo Brigato and Stavroula Mougiakakou**

ARTORG Center, Graduate School for Cellular and Biomedical Sciences, University of Bern, Switzerland
`{name}.{lastname}@unibe.ch`
https://github.com/lubnaa25/MealSAM_food_annotator

**Abstract.** The advancement of artificial intelligence (AI) in food and nutrition research is hindered by a critical bottleneck: the lack of annotated food data. Despite the rise of highly efficient AI models designed for tasks such as food segmentation and classification, their practical application might necessitate proficiency in AI and machine learning principles, which can act as a challenge for non-AI experts in the field of nutritional sciences. Alternatively, it highlights the need to translate AI models into user-friendly tools that are accessible to all. To address this, we present a demo of a semi-automatic food image annotation tool leveraging the Segment Anything Model (SAM) [15]. The tool enables prompt-based food segmentation via user interactions, promoting user engagement and allowing them to further categorise food items within meal images and specify weight/volume if necessary. Additionally, we release a fine-tuned version of SAM's mask decoder, dubbed MealSAM, with the ViT-B backbone tailored specifically for food image segmentation. Our objective is not only to contribute to the field by encouraging participation, collaboration, and the gathering of more annotated food data but also to make AI technology available for a broader audience by translating AI into practical tools.

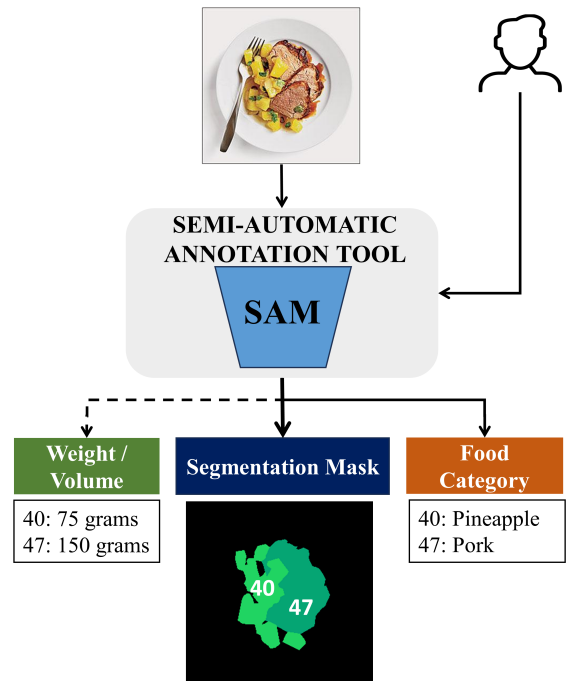

**Figure 1**: Inputs and outputs of the semi-automatic tool

## 1 Introduction

The realm of artificial intelligence (AI) has witnessed extraordinary growth [34], especially in healthcare, demonstrating its potential in disease prevention, diagnosis, and personalised management [12, 24]. Nutrition and diet-related diseases, crucial yet often overlooked, stand as key areas where AI can make a significant impact. This is especially pertinent amid the global increase in non-communicable diseases like obesity, diabetes, and cardiovascular diseases, highlighting the necessity for unbiased, accurate, and cost-efficient dietary intake monitoring and assessment [20, 31, 32]. While traditional dietary assessment methods have been the norm, they are time-consuming, expensive, and error-prone due to reliance on people's memory instead of objective measurements [29, 28]. Consequently, exploring alternatives such as visual estimations through computational image analysis is essential for improving the accuracy of dietary assessment across various socio-economic contexts [13, 5, 4, 14, 21].

Significant strides have been made towards leveraging AI and computer vision to automate dietary assessment through the analysis of meal images, achieving levels of accuracy comparable to those of dietitians [17, 23, 30]. As efforts continue to automate the dietary assessment process using AI, diverse models have emerged and undergone further exploration in food image analysis tasks of segmentation, classification, and volume estimation, opening up novel avenues for improved accuracy [15, 16, 22, 10, 35, 25, 8, 1]. However, the challenge of scarcity of annotated food data essential for model training and evaluation persists. Even though open-source food image datasets [3, 27, 2, 1] serve as valuable resources for food segmentation, classification, and volume estimation, they often lack thorough aggregated annotations. Moreover, there is an ongoing need to transform AI models into practical tools accessible to a wider audience, especially to non-AI experts [26, 9]. Collaboration between AI and non-AI experts, particularly nutrition specialists like dietitians, is crucial for advancing the field.

We acknowledge the labor-intensive nature of manually segmenting or annotating food items on a plate and, therefore, leverage AI's capabilities. Previous research has developed AI tools for image-based automatic segmentation, focusing primarily on medical images, as described in [19]. It has also been seen that past works have ventured into semi-automatic food segmentation tools [21], but none, to our best knowledge, have been made publicly accessible. Therefore, this work presents a practical food annotation tool to simplify the food annotation tasks, covering segmentation, classification of food items, and, optionally, volume/weight annotation. To promote effective user interaction, we leverage the capabilities of an open-source prompt-driven segmentation model known as the Segment Anything Model (SAM) [15]. SAM receives image inputs alongside prompts, which, in our tool, are user interactions via clicks within the meal image.

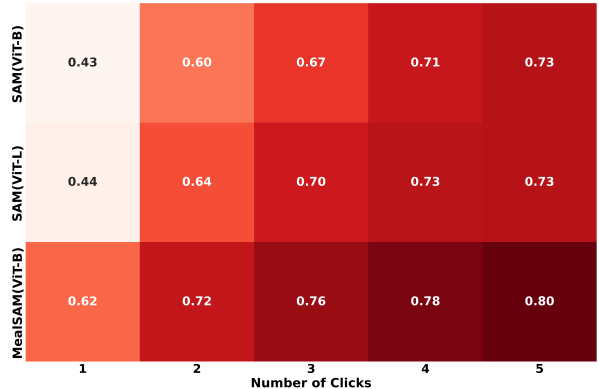

**Figure 2**: MealSAM vs SAM-variants on FoodSeg103 test set in terms of IoU (>=0.5) as a function of the number of clicks - average of three runs

Our primary contribution is an open-source tool, supporting not only pretrained versions of SAM but also MealSAM, which is a fine-tuned version of the SAM's mask decoder with the vision transformer base (ViT-B) backbone. As such, the tool is specifically designed to facilitate AI-driven food annotation. In this demo, we show the tool's capabilities by annotating meal images from the FoodSeg103 test set [33], covering segmentation, classification, and, optionally, weight/volume estimation. Our broader aim is to make AI accessible to a wider audience, including non-AI specialists, which is essential for advancing technology in nutrition research.

## 2 Materials and Methods

### 2.1 Prompt-based semi-automatic segmentation

SAM [15] is a notable advancement in image segmentation by supporting user prompts, showcasing remarkable zero-shot transfer skills across several domains [11, 18]. SAM is structured around three core components: an image encoder built on the vision transformer (ViT) architecture [7], a prompt encoder that accepts different input types such as points, boxes, text, and pre-existing masks, and a mask decoder that merges these inputs and produces a segmentation mask. The image encoder is available in different model scales, precisely, base (ViT-B), large (ViT-L), and huge (ViT-H).

In our annotation tool, prompts are equivalent to user clicks, thereby enhancing user engagement in image annotation. Our tool also allows users to select MealSAM or one of the three different pre-trained SAM model scales. Even though SAM can simultaneously generate multiple masks, we opt not to use this feature as we aim to create a singular mask for each food item for further annotation per food item. Each user interaction, through clicks, must target a specific food item, which is then processed to produce a single mask output. Users can mark desired areas (foreground) for masking with a left click, indicating the regions to be included. When using pre-trained variants of SAM, users can also denote unwanted areas (background) with a right click to mark exclusion from the mask. This approach is beneficial for accurately segmenting complex, layered food items. Upon a mask prediction, users are then prompted to categorise the segmented food/container item within the mask to complete the annotation process.

### 2.2 MealSAM

To enhance the segmentation mask quality, we opted to fine-tune the mask decoder specifically for the food domain. The training was done, with a batch size of 4, on a total of 18,320 binary food masks of the FoodSeg103 training set [33]. We used the Adam optimizer, with an initial learning rate of 1e-4 set to change every 10 epochs with loss as in [6]. To obtain prompts, we randomly sampled 1-5 points from the ground truth masks, which served as input points. For the evaluation of the test set, which contains 7,697 binary food masks, we again randomly sample between 1-5 points based on the ground truth masks, which are fixed while evaluating. This evaluation compares MealSAM against pre-trained SAM variants: ViT-B and ViT-L. The average intersection over union (IoU) across three different runs with 1-5 clicks, using a threshold of 0.5, as a function of the number of clicks, is presented in Figure 2. For both MealSAM and ViT-B, the average time taken for the generation of 1 mask on GPU acceleration (RTX A6000) was 0.23 seconds, while for ViT-L, it was 0.37 seconds. MealSAM not only achieves a higher average IoU than ViT-B but also surpasses ViT-L for food image segmentation while being faster, making it the preferred choice.

### 2.3 Tool design and implementation

The tool, developed in Python and utilising libraries and frameworks for image processing, is structured around three key components: segmentation, category selection, and weight or volume assignment to streamline the annotation process. We employ the Tkinter library for the GUI, PIL for image manipulation, OpenCV for image processing, and PyTorch for leveraging the deep learning back-end. We show an overview of the code architecture in the following snippet.

```
                    ── Tool Pseudocode ──
def ImageAnnotation(image, cat_file):
    # Input: Image file and category
    # Output: Annotated image,
    ↪    volume/weight of items (optional)
    UserUploadsImage(image)
    UserSelectsModel("SAM")   # MealSAM as
    ↪    default, user can choose
    while True:
        UserClicksOnFoodItem()   #
        ↪    Left-click to include in mask,
        ↪    right-click to exclude
        MaskPredictionbySAM()
        if MaskNotAccurate():
            if UserClearsClickedPoints():
                continue
            else:
                CorrectMaskWithBrush()
        UserSelectsFoodItem()
        Optional: UserInputsWeightVolume()
        UserValidatesAnnotation()
        if not MoreFoodItems():
            break
    UserClicksSave()
    SaveAnnotatedImage()
```

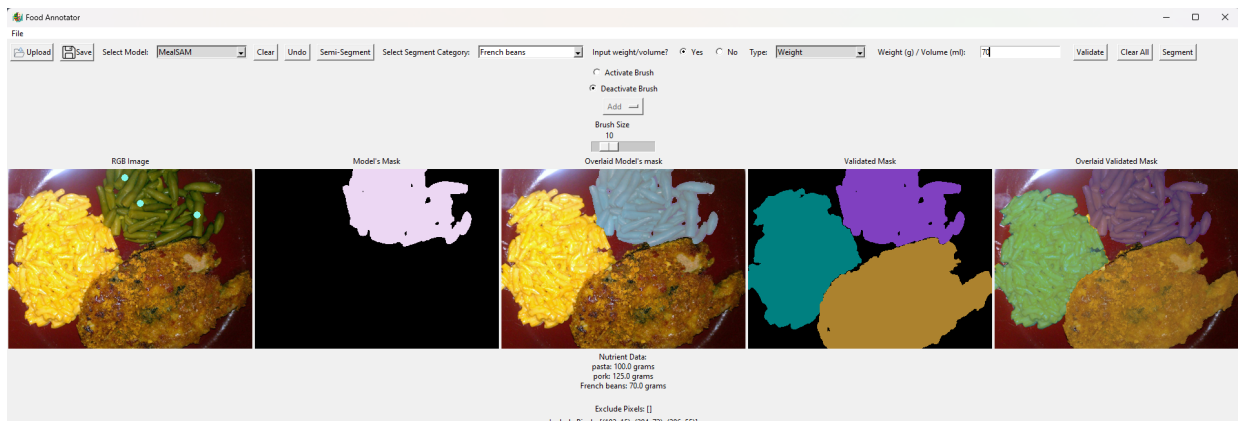

Figure 3: Illustration of meal image segmentation with annotated labels and volume estimation via semi-automatic tool

### 2.4 Dataset and food categories

For the demo, we make use of the food categories within Foodseg103. In the demo video, we annotate 2 meal images from the test set of Foodseg103 to demonstrate the capabilities of our segmentation tool. Annotations were made according to the ground truth classes specified by the dataset annotators.

### 2.5 Weight/Volume annotation

The tool incorporates an optional feature for annotating the weight or volume of segmented items motivated by the importance of quantitative analysis in food studies. After the segmentation step, the user optionally performs the annotation of weight/volume for a specific item, and a category is assigned. This part of the tool is specifically designed for cases where users are aware of ground truth weight/volume either through food weighing or volume computation.

## 3 Demo

In Figure 3, we show a comprehensively annotated meal image featuring segmentation, food labeling, and food weight/volume annotations. For this example, the operating machine specifics and segmentation tool configuration are as follows:

- Machine: Lenovo Legion 5 (15" AMD)
- Operating System: Windows
- Acceleration - GPU: NVIDIA RTX 3060
- Model: MealSAM
- Time Taken for complete annotation: 32.93 seconds

To begin using the segmentation tool, users upload their meal image. Based on the users' use case and accuracy needs, as previously mentioned, we allow them to choose between MealSAM and the different model sizes of the SAM image encoder. To perform the segmentation, the user first selects specific image areas by left-clicking, which the tool marks with blue dots as seen in Figure 3. The tool also displays the exact coordinates of the included clicks. Users have the option to undo their last action in case any mistakes have occurred. Moving forward, by selecting `Semi-Segment`, the tool generates a predictive mask by utilising the specified inclusions and exclusions and shows it in the `Model's Mask` window. For enhanced visualisation, the generated mask is overlaid on the original image in `Overlaid Model's Mask`. In case the predicted mask is unsatisfactory, an additional brush feature is provided to correct the mask through the "Activate Brush" feature. If the user approves of the outcome, they can then categorise the segment. In case the item is not included in the dataset, the user has additionally the option to add a new category. We assign each category a specific color code to facilitate distinct visual differentiation. For those wishing to annotate volume or weight, in case of presence of ground-truth weight/volume, the tool allows users to assign a value and further validate the mask. The user may repeat this annotation cycle as needed until all food items have been annotated. An example of the meal annotation process can be viewed in the demo video here.

## 4 Conclusion

In this work, we present food image annotation tool that leverages a powerful prompt-based segmentation model (SAM) for seamless, user-friendly segmentation and categorisation, with optional weight/volume annotation of food items. We also release the fine-tuned version of the mask decoder for the base architecture of SAM, MealSAM, specifically designed for food image segmentation. By enabling prompt-based interactions, our tool not only facilitates the generation of accurately annotated food data while reducing the time involved in fully manually annotating food items in images but also encourages the involvement of non-AI specialist users in the process. Our results showcase the tool's efficacy in producing high-quality, detailed annotations, contributing significantly to the field of dietary assessment and offering a solution to the ongoing challenge of data scarcity. Future plans include supporting larger versions of MealSAM and allowing annotation at different levels of granularity depending on use cases. By making this tool open-source, we aim to foster a community of collaboration, inviting contributions that will further enhance its capabilities and increase the amount of data available in the food domain for training and evaluating AI models.

### Acknowledgements

This work was partly supported by the European Commission and the Swiss Confederation - State Secretariat for Education, Research and Innovation (SERI) within the project 101057730 Mobile Artificial Intelligence Solution for Diabetes Adaptive Care (MELISSA).